\documentclass[letterpaper]{article} 
\usepackage{aaai24}  

\usepackage{times}  
\usepackage{helvet}  
\usepackage{courier}  
\usepackage[hyphens]{url}  
\usepackage{graphicx} 
\usepackage{natbib}  
\usepackage{caption} 
\usepackage{algorithm}
\usepackage{algorithmic}

\usepackage{newfloat}
\usepackage{listings}
\DeclareCaptionStyle{ruled}{labelfont=normalfont,labelsep=colon,strut=off} 

\usepackage{epsfig}
\usepackage{amsmath}
\usepackage{amssymb}
\usepackage{amsfonts, bm}
\usepackage{booktabs}
\usepackage{tikz}
\usepackage{hhline}
\makeatletter
\newcommand{\shorteq}{%
  \settowidth{\@tempdima}{-}
  \resizebox{\@tempdima}{\height}{=}%
}
\makeatother
\usepackage[mode=buildnew]{standalone}
\usepackage{multirow}
\usepackage{array}
\usepackage{eucal}

\floatstyle{ruled}
\newfloat{listing}{tb}{lst}{}
\floatname{listing}{Listing}
\title{LatentSwap: An Efficient Latent Code Mapping Framework for Face Swapping}
\author{
    Changho Choi\textsuperscript{\rm 1}\thanks{Equal contribution.},
    Minho Kim\textsuperscript{\rm 2}\footnotemark[1],
    Junhyeok Lee\textsuperscript{\rm 3},
    Hyoung-Kyu Song\textsuperscript{\rm 4},
    Younggeun Kim\textsuperscript{\rm 5},
    Seungryong Kim\textsuperscript{\rm 1}\thanks{Corresponding author.}
}
\affiliations{
    \textsuperscript{\rm 1}Korea University\\
    \textsuperscript{\rm 2}Massachusetts Institute of Technology\\
    \textsuperscript{\rm 3} Supertone Inc. \\
    \textsuperscript{\rm 4} Nota Inc. \\
    \textsuperscript{\rm 5} Optimizer AI \\
    changho98@korea.ac.kr,
    lgmkim@mit.edu,
    jun.hyeok@supertone.ai,
    hyoungkyu.song@nota.ai,
    eyfydsyd97@snu.ac.kr,
    seungryong\_kim@korea.ac.kr,

}

\usepackage{bibentry}

\begin{document}

\maketitle

\begin{abstract}
We propose LatentSwap, a simple face swapping framework generating a face swap latent code of a given generator. Utilizing randomly sampled latent codes, our framework is light and does not require datasets besides employing the pre-trained models, with the training procedure also being fast and straightforward. The loss objective consists of only three terms, and can effectively control the face swap results between source and target images. By attaching a pre-trained GAN inversion model independent to the model and using the StyleGAN2 generator, our model produces photorealistic and high-resolution images comparable to other competitive face swap models. We show that our framework is applicable to other generators such as StyleNeRF, paving a way to 3D-aware face swapping and is also compatible with other downstream StyleGAN2 generator tasks. The source code and models can be found at \url{https://github.com/usingcolor/LatentSwap}.
\end{abstract}

\section{Introduction}
\label{sec:intro}

Face swapping aims to produce an image that has the identity of the source image
while maintaining the target image attributes (e.g., facial expression, background,
and pose). 

While pioneering approaches primarily used 3D priors~\cite{blanz2004exchanging, lin2012face, yang2019exposing}, applying GANs~\cite{goodfellow2020generative} has produced more realistic face swap images~\cite{petrov_2020, bao_2018, li_2020, chen_2020, wang_2021_hififace,kim_2022, zhu_2021, xu_2022}. These applications focus on preserving the source identity information by using independent identity embedders such as ArcFace~\cite{deng2019arcface}, or preserving the shape and other facial attributes by using 3D face models such as 3DMM~\cite{blanz1999morphable}. Albeit producing credible results, these models crucially suffer from the typical GAN's slow and unstable training. Furthermore, each model using different datasets for training makes reproducing relevant results difficult. To solve this problem, recent works proposed a smooth identity embedder~\cite{kim_2022}, or attempted to incorporate pre-trained StyleGAN2~\cite{karras_2020_stylegan2} weights to take advantage of StyleGAN2's stability and performance. In a new direction, there also have been attempts to use diffusion probability models to perform face swapping~\cite{kim2022diffface}.


To utilize the stability and performance of StyleGAN2 generators into face swapping task, recent works~\cite{zhu_2021, xu_2022, xu2022styleswap, xu2022high} focused on translating images to relevant latent codes that can be piped into the StyleGAN2 generator, which typically employ identity encoders inspired by GAN inversion techniques. While these models achieved some success, they tend to require additional auxiliary modules such as face segmentation models or inputs such as face parsing models. Consequently, such networks do not exploit the full potential of the StyleGAN2 latent space.

In this paper, we propose an efficient face swapping model, where combined with a pre-trained GAN inversion model can produce face swap images given arbitrary source and target images. With a simple module dubbed the \textit{latent mixer} consisted of $5$ fully-connected layers and a residual connection, our model benefits from short and stable training. Furthermore, our model only takes randomly sampled source and target generator latent code pairs for training, and does not require a dataset besides using the information from pretrained models. For optimal performance we attached our model with PTI~\cite{roich2021pivotal} and the StyleGAN2 generator, benefitting from the full advantages of the StyleGAN2 latent space. We show that controllable generation between source and target is possible by systematically modifying loss objectives, and that specific attributes of the face swapping are disentangled to particular layers of the latent code. Our framework is applicable to other generators such as StyleNeRF~\cite{gu2021stylenerf} and can be used concurrently with other downstream tasks such as editing specific attributes. The resultant face swap results are high quality and realistic, comparable to state-of-the-art models.

Our contribution can be summarized as follows:
\begin{itemize}
\item We propose a simple and fast-training framework called the \textit{latent mixer} to generate face swap latent codes for a generator, which does not require additional
dataset for the purpose to train our model.

\item While fully exploiting the applied generator's latent space, our module can generate face swap images for real, wild images by utilizing an arbitrary pre-trained GAN Inversion network compatible with the generator.

\item Our framework allows for controlling generation by tweaking coefficients of three simple loss functions.

\item Our model is applicable to downstream tasks such as editing and to other generators such as StyleNeRF, opening up possible applications towards 3D face swapping.
\end{itemize}

\section{Related Work}
\label{sec:relatedworks}

\paragraph{GAN Inversion.}

GAN Inversion is a technique to map a real image to a generator space latent code, which can then be further used for generator-related downstream tasks such as editing~\cite{harkonen_2020, voynov_2020}. Naturally, after the advent of StyleGAN2, many GAN Inversion frameworks mapping real images onto the FFHQ domain space have been developed. 

The first approaches utilized optimization, where Image2StyleGAN~\cite{abdal_2019} and Image2StyleGAN++~\cite{abdal_2020} directly minimized the reconstruction distance from optimizing the latent codes, which subsequently led to the introduction of the $\mathcal{W+}$ space as well. These approaches achieve acceptable quality inversion results but in general take long times to generate such reconstructed images. 

Later on, learning-based methods~\cite{tov2021designing, richardson2021encoding, alaluf2021restyle, alaluf_2022} incorporated encoder networks to map the input image into latent space which can speed up the training and inference procedures significantly. However, compared to optimization-based methods it does not directly manipulate the latent code, hence lacks details and often requires a further optimization process. For example, PTI~\cite{roich2021pivotal} optimizes the generator parameters after using e4e~\cite{tov2021designing} to find the $\mathcal{W+}$ latent space latent code which leads to better visual quality.

\paragraph{Latent Space Manipulation.}
\label{sec:latentmanip}
Latent space editing uses a learned manifold of GAN~\cite{goodfellow2020generative} to control the image attributes such as age, pose, and expression.

Latent space editing methods can employ supervision, namely labeled datasets or pre-trained classifiers, to find the directions on the latent space which corresponds to the attributes~\cite{shen_2020, abdal_2021, patashnik_2021}. InterFaceGAN~\cite{shen_2020} disentangled the latent space employing linear transformations, while StyleFlow~\cite{abdal_2021} adopted a normalizing flow model. StyleClip~\cite{patashnik_2021} uses CLIP~\cite{radford2021learning} guided loss to edit via a text prompt. To the contrary, unsupervised methods search the directions first and analyze which attributes are associated with them~\cite{harkonen_2020, shen2021closed, voynov_2020, wang_2021_gan_geometric}. Examples include GANSpace~\cite{harkonen_2020} which performed PCA on latent codes, SeFa~\cite{shen2021closed} which used eigenvector decomposition, and Voynov et al.\cite{voynov_2020} which utilized mutual information. 
\section{Methodology}
\label{sec:methods}

\begin{figure*}
    \centering
    \setlength{\tabcolsep}{1pt}
    \includestandalone[width=0.80\textwidth]{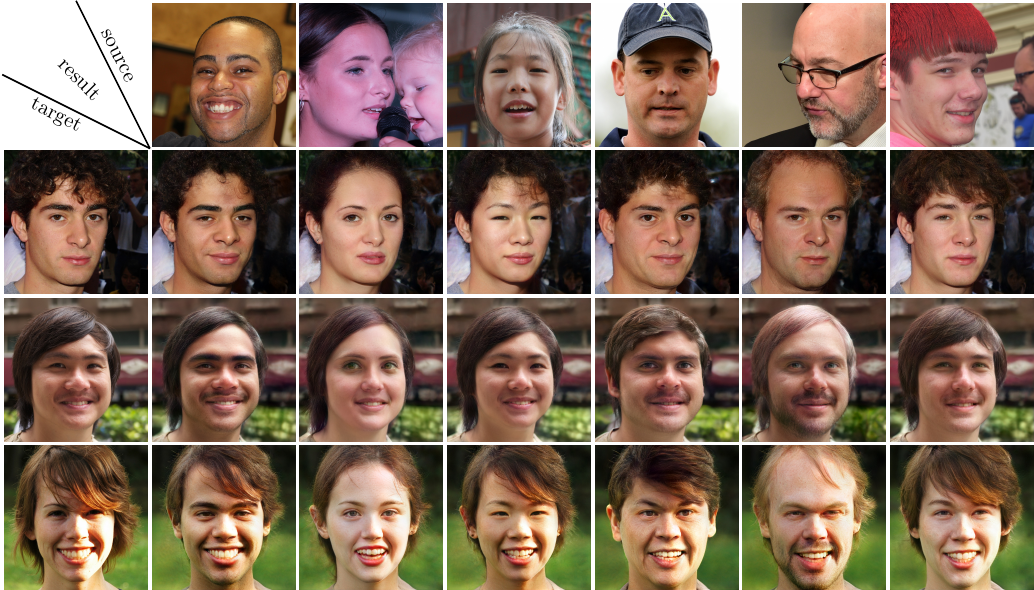}
   
    \caption{\textbf{Face swapping results by LatentSwap.} The model, given a source and a target image, replaces the identity of the target image (each row) to the identity of the source image (each column) while maintaining the attributes of the target images such as background and lighting. More results can be found in supplementary material.}
    \label{fig:teaser}
    \vspace{-2mm}
\end{figure*}

Given a generator trained on a face dataset, we attach a simple yet effective module to combine source and target latent codes to perform face swapping. Unlike previous methodologies~\cite{DeepFakesHttpsGithub2021, bao_2018, chen_2020, li_2020, wang_2021_hififace, xu_2022, zhu_2021, kim_2022}, we do not require any dataset for training. During training, the source/target latent codes are randomly sampled, and face swapping is performed using only these two latent codes as input. For inference, we apply a pre-trained GAN inversion model to map the real images onto the generator latent space. Besides the pre-trained inversion model, the generator, and the identity embedder~\cite{kim_2022} just to compute the training loss, no other pre-trained models have been used. We use the StyleGAN2~\cite{karras_2020_stylegan2} generator and work on its latent space, due to its versatility and good generation quality. 

\vspace{3pt}\noindent\textbf{Notations.}
On an arbitrary space $\mathcal{A}$, we denote the source and target latent codes and the swapped latent code formed by the aforementioned source and target respectively as $l_{\text{s}}^\mathcal{A}$, $l_{\text{t}}^\mathcal{A}$, and $l^\mathcal{A}_{\text{swap}}$. The latent mixer $\mathcal{F}^{\mathcal{A}}$ generates the swapped latent code $l_{\text{swap}}^{\mathcal{A}} = \mathcal{F}^{\mathcal{A}}(l_{\text{s}}^\mathcal{A}$, $l_{\text{t}}^\mathcal{A})$. The generator $G$ produces the corresponding image $I$ given any $18\times512$ dimensional vector: $I = G(l),\, l \in \mathbb{R}^{18\times512}$. We drop the superscript $\mathcal{W+}$ for swapped latent codes produced from $\mathcal{W+}$ latent space, unless distinctions are required.

\subsection{Latent Mixer Architecture}
\label{sec:model}
The goal of the latent mixer is to allow the swapped code to sufficiently capture the source’s identity without letting it deviate significantly from the target’s attributes with small computational resources.

\begin{figure*}[!ht]
    \centering
    \setlength{\tabcolsep}{1pt}
    
    \includestandalone[width=0.98\textwidth]{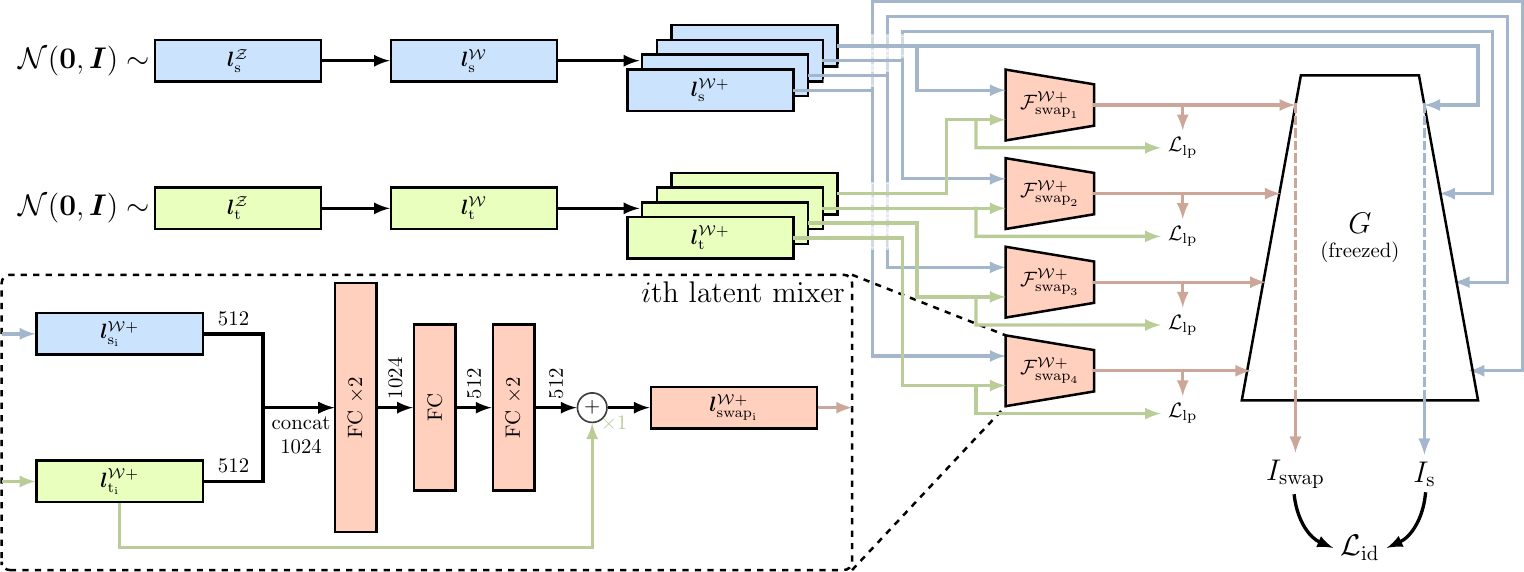}
    
    \caption{\textbf{Overall training scheme of the LatentSwap model.} We sample source and target latent codes randomly from a normal distribution, which are mapped onto $\mathcal{W}$ and subsequently copied 18-fold as a vector on $\mathcal{W+}$ space. The $\mathcal{W+}$ space vector is piped into the latent mixer (whose detailed structure is in the black dotted box). The swapped latent codes are then fed into the generator (the StyleGAN2 pre-trained weights) at different layers to generate the final face-swapped image. The gradient can still flow through the generator, but it remains freezed throughout training.}
    \label{fig:training}
    \vspace{-5mm}
\end{figure*}

Each latent mixer consists of $5$ fully-connected layers with Swish~\cite{ramachandran2017searching} activation. The latent mixer takes the source and the target latent codes as input. The two latent codes are then concatenated, and it goes through $5$ fully-connected layers in sequence. This output is then added to the target latent code to produce the final swap latent code. Each latent mixer operates on each of the 18 layers of the latent codes on $\mathcal{W+}$ space. We add the target latent code to the output of the last fully-connected layer~\cite{he2016deep} to generate the final swapped latent code $l_{\text{swap}}$. It is possible to control the residual connection coefficient, \textit{i.e.}, how much $l_{\text{t}_{\text{i}}}^{\mathcal{W+}}$ to add: see Appendix 1 for details. Detailed structure of the latent mixer is illustrated in Figure~\ref{fig:training}. 

While we can generate quality face swapped images by only using $2$ or $3$ layers, we note that the optimal performance between training speed and performance is at $5$ fc layers. We further discuss this in supplementary information by plotting the loss function during training and the performance of the network for $2$, $3$, $5$, and $7$ layers, which shows that quality improvement decreases as layers increase, while the increasing network size burdens the training speed.

\subsection{Training Procedure}
\label{sec:training}

First, we sample the latent code for the source and target from a standard normal distribution: $l_{\text{s}}^{\mathcal{Z}}, l_{\text{t}}^{\mathcal{Z}} \sim \mathcal{Z} = \mathcal{N}(\bm{0}, \bm{I})$, where $l_{\text{s}}^{\mathcal{Z}}, l_{\text{t}}^{\mathcal{Z}} \in \mathbb{R}^{512}$. We transform this to $\mathcal{W}$ space via 8 pre-trained fc layers, as used in StyleGAN2, which are also $512$-dimensional: $l_{\text{s}}^{\mathcal{W}}, l_{\text{t}}^{\mathcal{W}} \in \mathcal{W} \subset \mathbb{R}^{512}$. This latent code is then duplicated $18$ times, resulting in a latent code on the $\mathcal{W+}$ space consisting of $18$ layers, with each layer consisting of an \textit{identical} $512$-dimensional vector: a $\mathcal{W+}$ space latent code therefore has dimension $18 \times 512$. We note that the general $\mathcal{W+}$ space is larger, which is a set of any concatenation of 18 arbitrary latent codes on the $\mathcal{W}$ space~\cite{abdal_2019}. Note for inference, PTI directly produces the $\mathcal{W}$-space latent codes.

Each layer of the W+ space is followed by a latent mixer, which produces a layer-wise swapped latent code. We then collect these layer-wise codes from all the $18$ latent mixers and concatenate them to obtain the full $18 \times 512$-dimensional swapped face latent code $l^\mathcal{W+}_{\text{swap}}$. 

After obtaining the face swap latent code, we calculate the training loss and backpropagate the gradients through the model $G$. We only update the weights in the latent mixer using the optimizer while keeping the StyleGAN2 weights constant. The gradient flows back to each of the latent mixers at different points of the generator, achieving layer-wise distinction that corresponds to different resolutions of the image~\cite{karras_2019}. Figure~\ref{fig:training} denotes this mechanism graphically.

\subsection{Loss Functions}
\label{sec:loss}

We use three simple loss functions to train our model, consisting of the identity loss $\mathcal{L}_{\text{id}}$, latent penalty loss $\mathcal{L}_{\text{lp}}$, and the shape loss $\mathcal{L}_{\text{s}}$ added together:
\begin{equation}
    \mathcal{L} = \mathcal{L}_{\text{id}} + \lambda \mathcal{L}_{\text{lp}} + \mu \mathcal{L}_{\text{s}}. 
    \label{eq:gloss}
\end{equation}
where we set $\lambda=10^2$ as the default value and fix $\mu = 0.1$, which is set empirically so that the shape loss term will be similar size to the other loss terms, indicating that all three loss terms will contribute during training.

\begin{figure}[t]
  \includestandalone[width=0.47\textwidth]{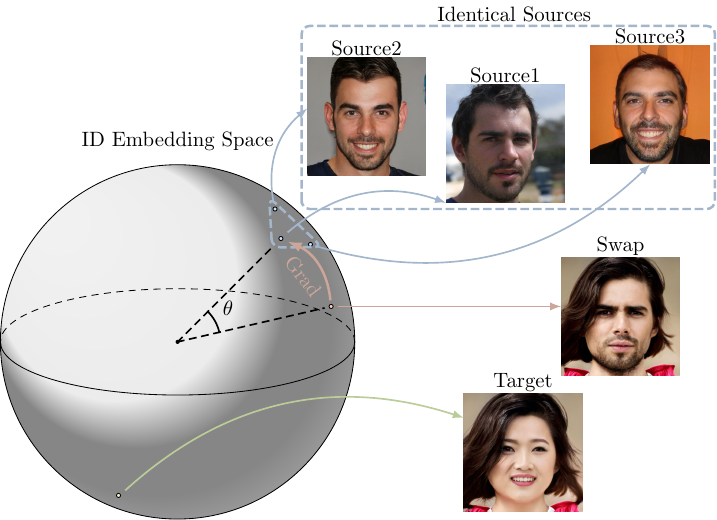}
  \caption{\textbf{Schematic description of the ID loss.} The source and swapped images are mapped to the Smooth-Swap~\cite{kim_2022} identity embedding space. The loss objective is to minimize the cosine distance between the identity embeddings.}
  \label{fig:idloss}
  \vspace{-5mm}
\end{figure}

\vspace{3pt}\noindent\textbf{Identity Loss.}
The identity loss $\mathcal{L}_{\text{id}}$ (ID loss) is defined following the typical conventions of general face swapping models~\cite{li_2020, chen_2020, zhu_2020, wang_2021_hififace, xu_2021, kim_2022} as the cosine distance between the identity embeddings of the source and swapped images. Therefore, the ID loss helps the swapped face to share the same identity as the source. This can be written as:
\begin{equation}
    \mathcal{L}_{\text{id}} = 1-\cos(\textrm{ID}(I_{\text{s}}), \textrm{ID}(I_{\text{swap}})). \label{eq:idloss}
\end{equation}
where $\textrm{ID}: \mathbb{R}^{3\times H \times W} \rightarrow \mathbb{R}^{512}$ denotes the identity embedder and $I_{\text{s}}$, $I_{\text{swap}}$ denotes the source and swapped images. This is described schematically in Figure~\ref{fig:idloss}.

We use the identity embedder from Smooth-Swap, which has a smooth embedding space leading to more stable and fast training \footnote{A detailed discussion on the performance of the Smooth-Swap embedder compared to ArcFace is provided on the original Smooth-Swap paper.}. We implemented the identity embedder ourselves since the official implementation is not released.

\vspace{3pt}\noindent\textbf{Latent Penalty Loss.}
The equally important aspect in face swapping tasks alongside identity preservation is to preserve target attributes. Provided that the mappings from one latent space to another is continuous on the StyleGAN domain space, the ideal face swap and the target latent codes should be close in the latent space, especially considering that all attributes of the target should ideally be preserved after face swapping~\cite{patashnik_2021}. This does not necessarily mean that the $\mathcal{F}^{\mathcal{W+}}$ map will be able to generate a latent code which corresponds to the swapped face (see results for $\mathcal{Z}$ in Figure~\ref{fig:domainspace} and Appendix 2.). 

We define the latent penalty loss $\mathcal{L}_{lp}$ at a given latent space as the mean squared error (MSE) distance on the $\mathcal{W+}$ space between the target and swapped latent codes:
\begin{equation}
\label{eq:lploss}
    \mathcal{L}_{\text{lp}} = \frac{1}{\mathrm{dim}(l_{\text{t}}^{\mathcal{W+}})} \lVert l_{\text{t}}^{\mathcal{W+}} - l^{\mathcal{W+}}_{\text{swap}} \rVert_2^2.
\end{equation}

\vspace{3pt}\noindent\textbf{Shape Loss.} Shape loss utilizes 2D landmark keypoints as
geometric supervision to accurately guide the model to constrain the face shape. Following previous models~\cite{wang_2021_hififace}, we extract the 3DMM~\cite{blanz1999morphable} coefficients for the source and target images, and construct the fused 3DMM coefficient by concatenating the source image's identity and the target image's expression and posture to generate the fused image facial landmarks.

We define the shape loss to be the $L_1$ distance between the 3D face landmarks of the fused facial landmark $q_{\text{fuse}}$ and the 3D face landmarks of the swapped image $q_{\text{swap}}$, which could be written as

\begin{equation}
    \mathcal{L}_{\text{s}}=\lVert q_{\text{swap}}-q_{\text{fuse}} \rVert_1.
\end{equation}

\subsection{Incorporating Real Images}

The LatentSwap framework operates on the generator latent space. Therefore, we require a GAN Inversion network to map the real image to the latent space. For this, we use PTI~\cite{roich2021pivotal} on optimization mode to map the real image to the $\mathcal{W}$ space latent code. While it is possible to directly map onto $\mathcal{W+}$ space using e4e~\cite{tov_2021} as the encoder network, we note that during the training process we opted to copy the $\mathcal{W}$ space latent codes to create a vector on $\mathcal{W+}$ space, and directly mapping onto $\mathcal{W+}$ space breaks this symmetry. After PTI generates the corresponding $\mathcal{W}$ space latent codes for the source and target images, we then optimize the StyleGAN2 weight using the generated $\mathcal{W}$ space source and target latent codes. The optimized weight is then used to generate the final face swap image.
\section{Experiments}
\label{sec:experiments}

\subsection{Implementation Details}
\label{sec:implementation}

We use the FFHQ pre-trained StyleGAN2~\cite{karras_2020_stylegan2} weights as our generator, which produces $1024\times1024$ images. We first analyze the results when the latent mixer takes $\mathcal{W+}$ space vectors. Subsequently, we study the effects of latent spaces, namely $\mathcal{Z}/\mathcal{W}$, and $\lambda$. We then perform layer-wise analysis by applying swapped latent codes at specific resolutions and apply editing techniques.

The batch size is 64 source-target pairs, of which there is a $1/8$ probability that the pair consists of the same latent codes for each source-target pair\footnote{Not to be interpreted as $8$ pairs within a batch of $64$ are selected to have the same source and target.}. We follow the procedure outlined in Smooth-Swap\cite{kim_2022} to train the identity embedder, which is used to calculate the ID loss $\mathcal{L}_{\text{id}}$. We train for $200$k steps with the AdamW~\cite{loshchilov2017decoupled} optimizer and a learning rate of $lr=0.0001$ on $2$ NVIDIA A$100$ $80$GB GPUs on a DGX Station. For our default settings using $18$ latent mixers, training takes approximately $77$ hours. The average inference time is 98s, with most of the time spent on the additional optimization steps; going through the fc layers only
takes 0.2s. See Table~\ref{tab:quant_} for number of parameter comparison.

\subsection{Experimental Results}
\label{sec:results}
 
\begin{table}[t]
    \centering
    \resizebox{\linewidth}{!}{
    \begin{tabular}{c@{}ccccc}
    \toprule 
    Space & $\lambda$ & ID$\downarrow$ & Expression$\downarrow$ & Pose$\downarrow$ & FID$\downarrow$ \\ [0.5ex] 
    \midrule 
    $\mathcal{Z}$ & $10^{0\phantom{-}}$ & 0.469\phantom{0} & 0.158\phantom{00} & \phantom{0}52.0\phantom{00} &  \phantom{0}5.68\\ 
    \hline
    $\mathcal{Z}$ & $10^{1\phantom{-}}$ & 0.510\phantom{0} & 0.0828\phantom{0} & \phantom{0}18.2\phantom{00} &  \phantom{0}4.05\\ 
    \hline
    $\mathcal{Z}$ & $10^{2\phantom{-}}$ & 0.579\phantom{0} & 0.00775 & \phantom{00}1.36\phantom{0} &  \phantom{0}3.03\\ 
    \midrule
    $\mathcal{W}$ & $10^{0\phantom{-}}$ & 0.287\phantom{0} & 0.159\phantom{00} & \phantom{0}37.6\phantom{00} & 37.5\phantom{0} \\
    \hline
    $\mathcal{W}$ & $10^{1\phantom{-}}$ & 0.355\phantom{0} & 0.0919\phantom{0} & \phantom{0}12.9\phantom{00} & \phantom{0}8.57 \\
    \hline
    $\mathcal{W}$ & $10^{2\phantom{-}}$ & 0.529\phantom{0} & 0.0145\phantom{0} & \phantom{00}1.16\phantom{0} & \phantom{0}2.97 \\
    \hline \hline
    $\mathcal{W+}$ & $\phantom{1}0^{\phantom{-0}}$ & 0.469\phantom{0} & 0.191\phantom{00} & 190\phantom{.000} & 171\phantom{.000}\\
    \hline
    $\mathcal{W+}$ & $10^{-2}$ & 0.0492 & 0.242\phantom{00} & 288\phantom{.000} & \phantom{0}5.53 \\
    \hline
    $\mathcal{W+}$ & $10^{-1}$ & 0.0563 & 0.239\phantom{00} & 275\phantom{.000} & \phantom{0}4.96 \\
    \hline
    $\mathcal{W+}$ & $10^{0\phantom{-}}$ & 0.215\phantom{0} & 0.187\phantom{00} & \phantom{0}68.3\phantom{00} & 30.8\phantom{0} \\
    \hline
    $\mathcal{W+}$ & $10^{1\phantom{-}}$ & 0.315\phantom{0} & 0.159\phantom{00} & \phantom{0}23.7\phantom{00} & 19.7\phantom{0} \\
    \hline
    \boldmath$\mathcal{W+}$ & \boldmath$10^{2\phantom{-}}$ & 0.384\phantom{0} & 0.0847\phantom{0} & \phantom{00}7.55\phantom{0} & \phantom{0}6.24 \\
    \hline
    $\mathcal{W+}$ & $10^{3\phantom{-}}$ & 0.541\phantom{0} & 0.0108\phantom{0} & \phantom{0}0.802 & \phantom{0}2.95 \\
    \bottomrule 
\end{tabular}}
    \caption{\textbf{Metric performance of LatentSwap, for latent codes sampled from various latent spaces and with various $\lambda$ coefficients.} The result for the default configuration, taking $\mathcal{W+}$ space latent codes and training with $\lambda=10^2$, has been boldfaced.}
    \label{tab:quant}
    \vspace{-2mm}
\end{table}

Overall, when the latent space is fixed to $\mathcal{W+}$, the metrics show a clear trend that swapped images are closer to the source at lower $\lambda$, and closer to the target at higher $\lambda$. We see a similar trend for latent mixer operating on $\mathcal{W}$ latent codes, but not for the $\mathcal{Z}$ counterpart. For the custom dataset, Table~\ref{tab:quant} shows the results for various sampling spaces and loss function coefficients. We compare with different face swapping models in Table~\ref{tab:quant_}, with real image metrics presented in supplementary information. 

\begin{table}[t]
    \centering
    \begin{adjustbox}{width=\columnwidth,center}
    \begin{tabular}{c@{}cccc}
    \toprule
    Model\phantom{00} & ID$\uparrow$ & Expression$\downarrow$ & Pose$\downarrow$ & Params$\downarrow$\\ [0.5ex] 
    \midrule 
    DeepFakes \phantom{00} & $88.39$ & $0.1705$ & $13.38$ & Unknown \\ %
    FaceShifter \phantom{00} & $90.68$ & $0.1223$ & $7.65$ & $250$M \\ %
    SimSwap \phantom{00} & $89.73$ & $0.0879$ & $5.82$ & $120$M\\  %
    HifiFace \phantom{00} & $98.48$ & NA & $7.89$ & $244$M\\  %
    MegaFS \phantom{00} & $90.83$ & $0.1348$ & $7.92$ & $338$M\\  %
    RAFSwap \phantom{00} & $96.70$ & $0.1312$ & $7.59$ & Unknown\\ %
    InfoSwap \phantom{00} & $\textbf{99.67}$ & $ 0.1427$ & $9.07$ & $251$M \\ %
    \hline
    \textbf{Ours}\phantom{00} & $93.36$ & $\textbf{0.0673}$ & $\textbf{4.06}$ & $\mathbf{87}$\textbf{M}\\ 
    \bottomrule 
\end{tabular}
\end{adjustbox}
\caption{\textbf{Metric comparisons of our model on the FaceForensics++ dataset.} Our model achieves comparable performances to other face swapping models for ID and better performance for all other metrics.}

\label{tab:quant_}
\vspace{-5mm}
\end{table}

We follow the convention from previous face swapping studies~\cite{zhu_2020, xu_2022}. The ID metric is the accuracy of a classification among swapped faces and all original FF++ faces. The Expression and Pose are the mean
square error between the expression features/pose information between target and swapped images.

The anomalously low performance of the $\lambda=0$ establish that $\mathcal{L}_{\text{id}}$ by itself is insufficient to constrain our model to the suitable latent space, hence that $\mathcal{L}_{\text{lp}}$ is needed. 

\vspace{3pt}\noindent\textbf{Qualitative Results.} Our method generates realistic and high resolution face swaps. This shows that the pre-trained PTI model~\cite{roich2021pivotal} works well with our framework and has mapped the arbitrary image onto the $\mathcal{W}$ space.

\begin{figure}[!t]
    \centering
    \setlength{\tabcolsep}{1pt}
    \newcolumntype{x}{>{\centering\arraybackslash\vspace{0pt}}m{0.10\linewidth}}
    \begin{tabular}{xxxxxxxxxx}
        \scalebox{.45}{Source} &
        \scalebox{.45}{Target} &
        \scalebox{.45}{Deepfakes} & 
        \scalebox{.45}{FaceSwap} & 
        \scalebox{.45}{FaceShifter} & 
        \scalebox{.45}{MegaFS} &
        \scalebox{.45}{HifiFace} & 
        \scalebox{.45}{InfoSwap} & 
        \scalebox{.5}{Ours}
    \end{tabular}
    
    \includestandalone[width=0.98\linewidth]{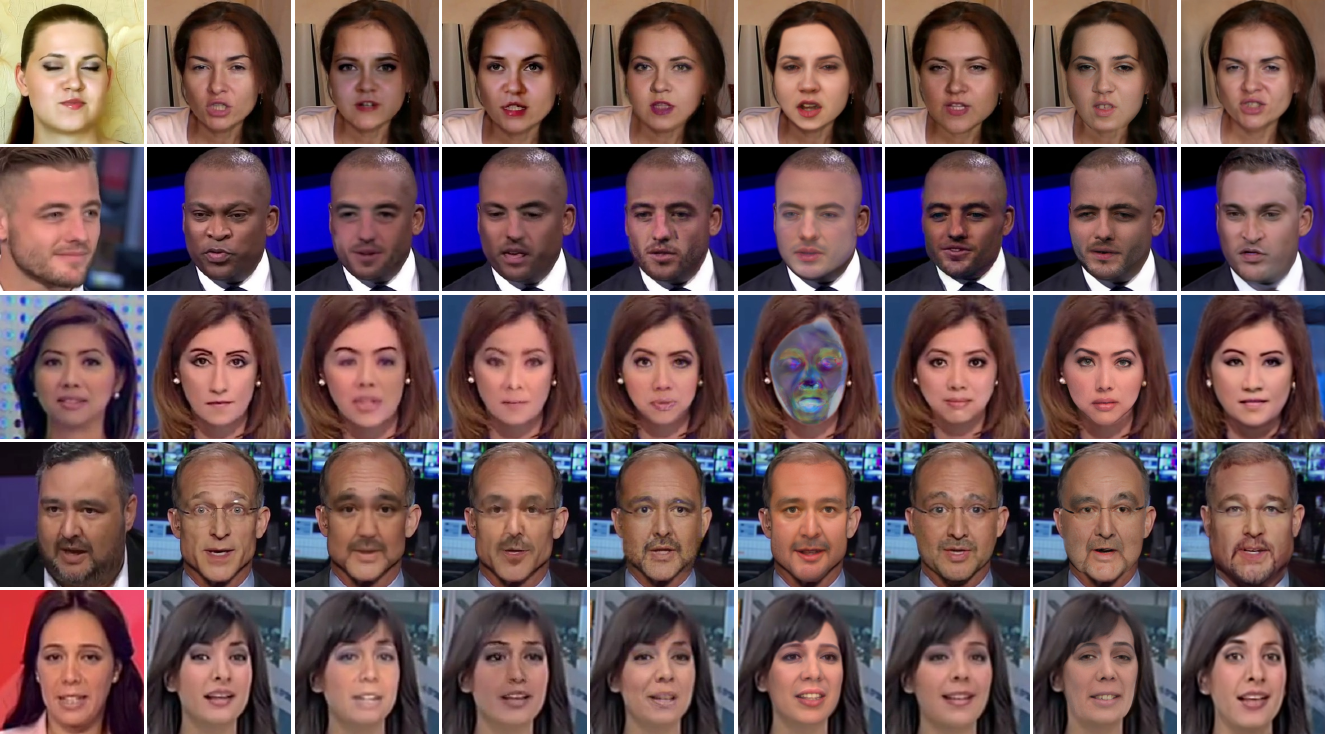}
    
      \caption{\textbf{Face swapping results of LatentSwap compared to other face swapping models on FF++.} We used the code and checkpoints from their official implementation. Our model performs source identity and target attribute preservation well compared to other models. 
      }
   \label{fig:quality}
   \vspace{-3mm}
\end{figure}

In general, the face shape of the swapped image closely resembles the target and does not get blurry when the face shape differs between source and target. Hairstyle, on the other hand, tends to follow the source image. Interestingly, this tendency has also been visible in Smooth-Swap, which may indicate that this phenomenon is linked to the  smooth identity embedder that both of the models use. 

These descriptions are more poignant when compared to conventional models such as SimSwap~\cite{chen_2020} and MegaFS~\cite{zhu_2021}. SimSwap gives an acceptable face swapping result, especially regarding background and lighting. However, it rarely changes the skin tone and is constrained to a $224 \times224$ image size. MegaFS performs qualitatively worse. This is primarily due to the segmentation model changing only specific parts of the target image, leading to unnatural background and being vulnerable to occlusion, alongside showing inexact lighting and face shape.

\subsection{Analysis and Ablation Study}
\label{sec:ablation}

\begin{figure}[t]
\centering
    \setlength{\tabcolsep}{0pt}
    \newcolumntype{x}{>{\centering\arraybackslash\vspace{0pt}}m{0.20\linewidth}}
    \begin{tabular}{xxxxx}
    \footnotesize{Source} &
    \footnotesize{Target} & 
    \footnotesize{$\mathcal{Z}$ Mixer} & 
    \footnotesize{$\mathcal{W}$ Mixer} & 
    \footnotesize{$\mathcal{W}+$ Mixer}
    \\
    \multicolumn{5}{c}{
    \includegraphics[width=\columnwidth]{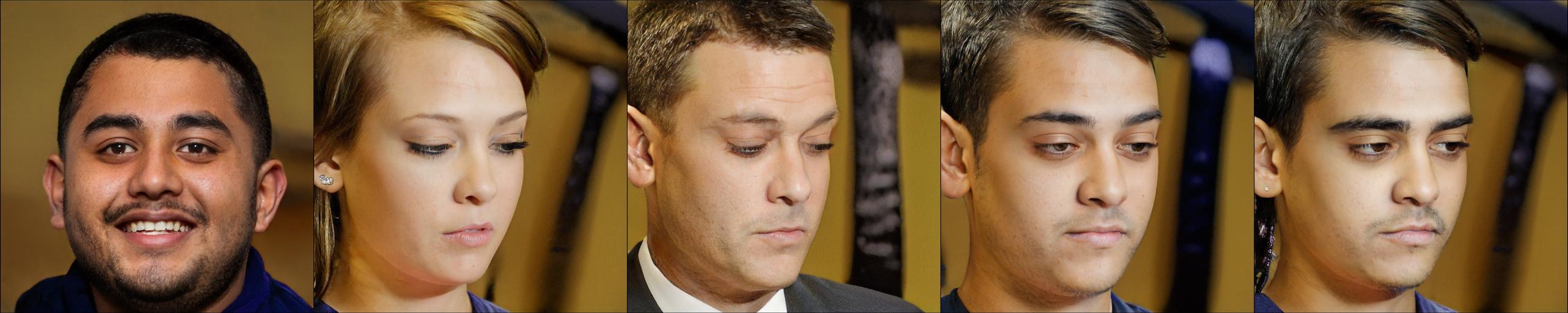}
    }
    \\[-2pt]
    \multicolumn{5}{c}{
    \includegraphics[width=\columnwidth]{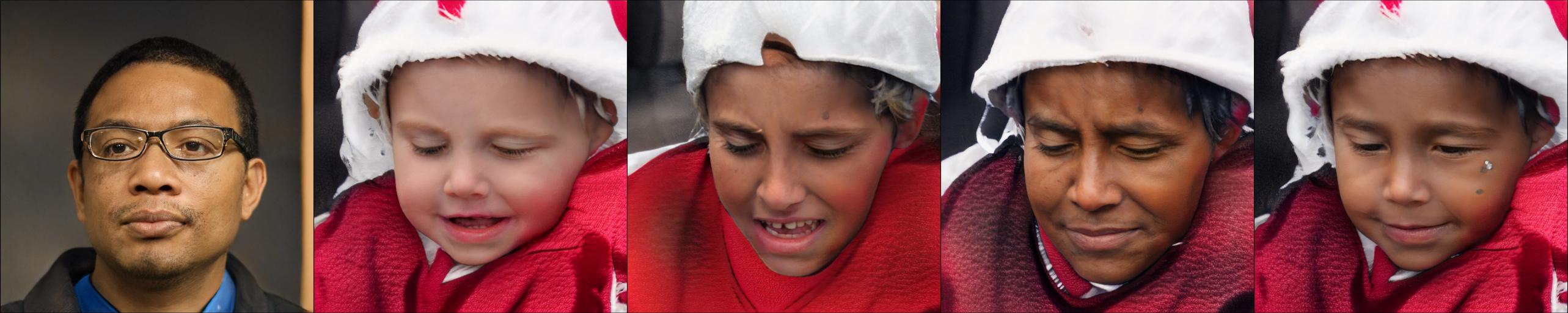}
    }
    \\[-2pt]
    \multicolumn{5}{c}{
    \includegraphics[width=\columnwidth]{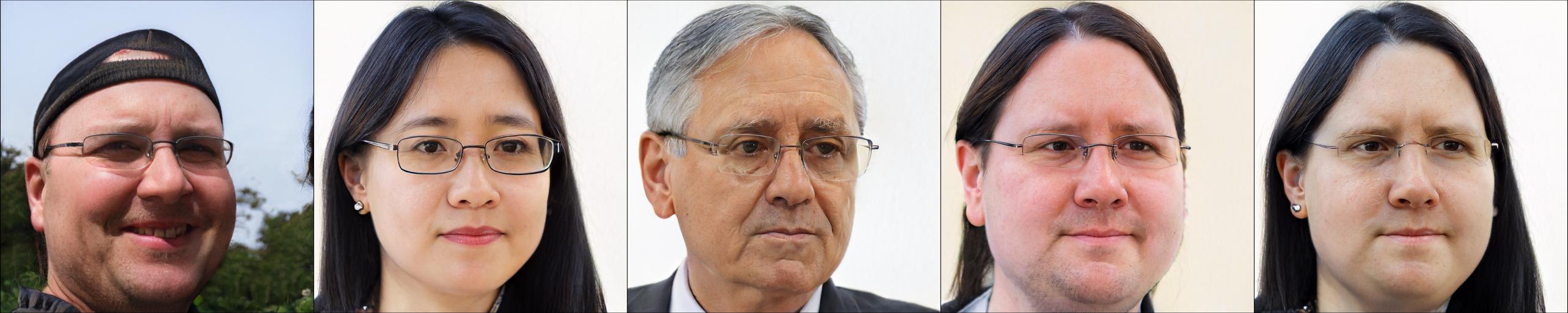}
    }
    \\
    \end{tabular}
\caption{\textbf{Performance of the latent mixer applied to latent codes from different latent spaces.} We used $\lambda = 10^1$ for $\mathcal{Z}$ and $\mathcal{W}$ space results, whereas for the $\mathcal{W}+$ space we used $\lambda = 10^2$. $\mathcal{W}+$ space results keep more target attributes compared to other latent spaces while maintaining source identity, which is not preserved for $\mathcal{Z}$ space results.}
\label{fig:domainspace}
\vspace{-5mm}
\end{figure}

\vspace{3pt}\noindent\textbf{Latent Space.} In this section, we apply the latent mixer to source/target latent codes on $\mathcal{Z}$ and $\mathcal{W}$, and compare them with $\mathcal{W+}$. The results are shown in Figure~\ref{fig:domainspace}.

\begin{figure}[t]
\centering
    \setlength{\tabcolsep}{0pt}
    \newcolumntype{x}{>{\centering\arraybackslash\vspace{0pt}}m{0.11\linewidth}}
    \begin{tabular}{xxxxxxxxx}
    \scriptsize{Source} &
    \scriptsize{Target} & 
    \scalebox{.66} {$\lambda=0$} & 
    \scalebox{.66} {$\lambda=10^{\text{-}2}$} & 
    \scalebox{.66} {$\lambda=10^{\text{-}1}$} & 
    \scalebox{.66} {$\lambda=10^{0}$} & 
    \scalebox{.66} {$\lambda=10^{1}$} & 
    \scalebox{.66} {$\lambda=10^{2}$} & 
    \scalebox{.66} {$\lambda=10^{3}$}
    \\
    \multicolumn{9}{c}{
    \includegraphics[width=\columnwidth]{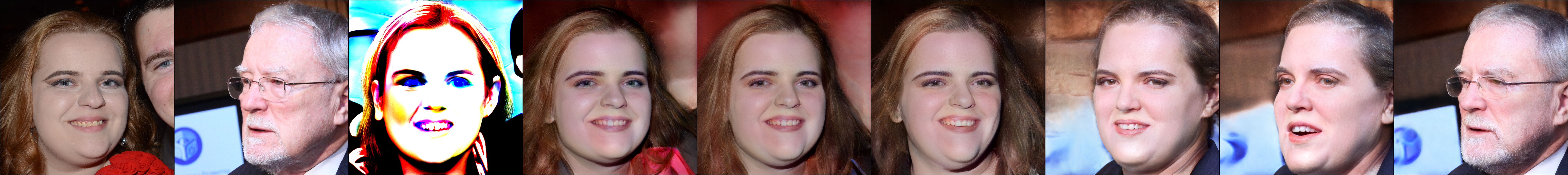}
    }
    \\[-2.5pt]
    \multicolumn{9}{c}{
    \includegraphics[width=\columnwidth]{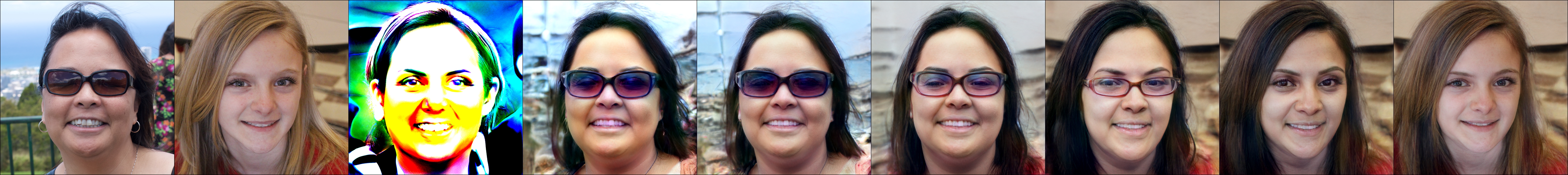}
    }
    \\[-2.5pt]
    \multicolumn{9}{c}{
    \includegraphics[width=\columnwidth]{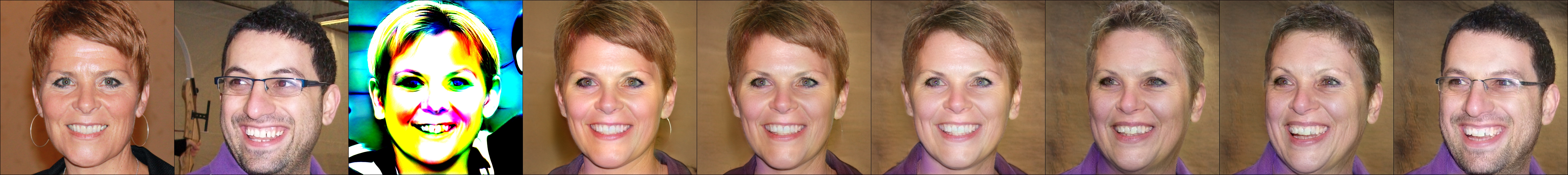}
    }
    \\[-2.5pt]
    \multicolumn{9}{c}{
    \includegraphics[width=\columnwidth]{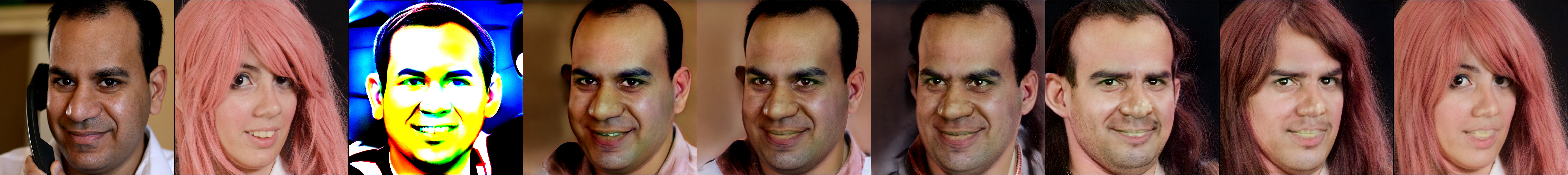}
    }
    \end{tabular}
\caption{\textbf{Comparison of face swapping results by various $\lambda$.} When $\lambda = 0$, the model trains only with the ID loss and outputs unrealistic images. As $\lambda$ is increased, we see a gradual shift of the face swap from source to target.}
\label{fig:lpablation}
\vspace{-6mm}
\end{figure}

Table~\ref{tab:quant} shows that taking inputs from $\mathcal{W+}$ and $\mathcal{W}$ space gives good and comparable results at their best-performing $\lambda$ values. However, due to $\mathcal{W}$ space being smaller than $\mathcal{W+}$ space, the images lack details, and with some target attributes such as hair and background distorted. Taking inputs from $\mathcal{Z}$ fails to find a suitable latent code corresponding to the swapped face throughout all $\lambda$ as seen in the high ID metric values for all $\mathcal{Z}$ space results in Table~\ref{tab:quant}.

A possible explanation comes from that the sample space of $\mathcal{Z}$ distribution is $\mathbb{R}^{512}$. Therefore, the latent mixer output, being in $\mathbb{R}^{512}$, also belongs to the sample space of $\mathcal{Z}$ and the latent mixer output can never escape the sample space of $\mathcal{Z}$. In contrast, $\mathcal{W}$ and $\mathcal{W+}$ are subspaces embedded in $\mathbb{R}^{512}$. Therefore, it is possible for the latent mixer to output latent codes \textit{outside} $\mathcal{W}$ or $\mathcal{W+}$ space, where it is more likely for the latent code for the swapped face to exist.

\begin{figure}[t]
    \centering
    \includegraphics[width=0.75\columnwidth]{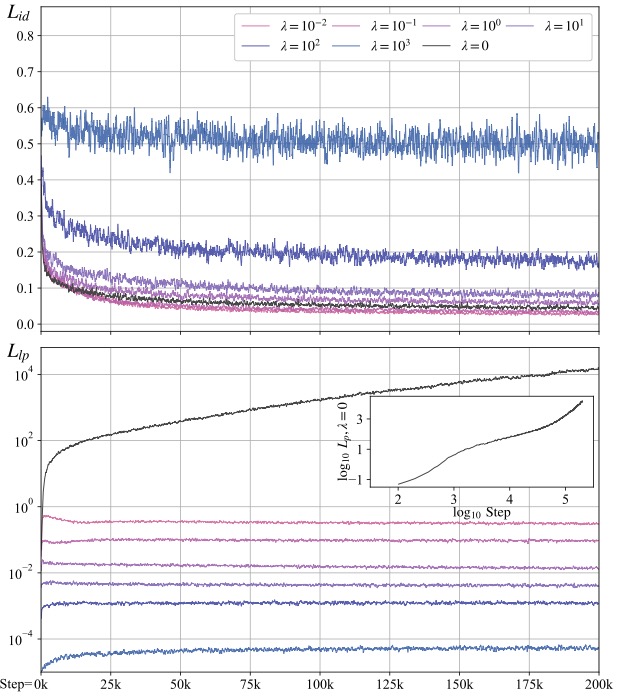}
    \caption{\textbf{$\mathcal{L}_{\text{id}}$ (top panel) and $\mathcal{L}_{\text{lp}}$ (bottom panel) by training step, for various $\lambda$.} The inset is the $\log-\log$ plot of number of steps and $\mathcal{L}_{\text{lp}}$. The losses are window averaged with a window width of $250$ steps. 
    }
    
    \label{fig:lplossgraph}
    \vspace{-6mm}
\end{figure}

\vspace{3pt}\noindent\textbf{Latent Penalty Loss.} We analyze the performance of various $\lambda$ values, while maintaining other parameters, which is presented in Figure~\ref{fig:lpablation}. We confirm visually that the result resembles the source image for small $\lambda$, and the target image for large $\lambda$, with a smooth transition in between. Intriguingly, the source identity is swapped at higher $\lambda$ values compared to other attributes such as background and lighting. When $\lambda=0$, the images are unrealistic and the model is clearly sampling outside the intended image domain. Figure~\ref{fig:lplossgraph} shows that (unused) $\mathcal{L}_{\text{lp}}$ diverges when $\lambda=0$.

This behaviour stems from $\mathcal{L}_{\text{lp}}$ providing a competing objective to $\mathcal{L}_{\text{id}}$, and training reaching equilibrium when the two loss terms are similar in magnitude. Therefore, $\lambda$ effectively limits the deviation of the swapped latent code from the target to $\mathcal{L}_{\text{lp}} \sim \frac{1}{\lambda} \: \mathcal{L}_{\text{id}}$, which can be seen in the bottom panel of Figure~\ref{fig:lplossgraph}.

For $\lambda=0$ we note an analogue between the behavior of $\mathcal{L}_{\text{lp}}$ and diffusion. After initial stabilization, the swapped latent code performs a random walk, corresponding to the classical diffusion formula $\mathcal{L}_{\text{lp}} \sim \textrm{step}$ at intermediate steps. 
Interestingly, at larger steps we observe superdiffusion $\mathcal{L}_{\text{lp}} \sim \textrm{step}^\nu, \: \nu>1$ 
While further work is needed to exactly account for this dynamical phase transition, this point could potentially be a indicator of the limit where StyleGAN2 latent space is well-behaved. 

\vspace{3pt}\noindent\textbf{Layer-wise Analysis.} In StyleGAN~\cite{karras_2019, karras_2020_stylegan2, karras_2021}, each of the 18 layers of the $\mathcal{W+}$ latent code is linked to a specific resolution, and hence to different semantics~\cite{karras_2019}. 

\begin{figure}[t]
\centering
    \setlength{\tabcolsep}{0pt}
    \newcolumntype{x}{>{\centering\arraybackslash\vspace{0pt}}m{0.143\linewidth}}
    \begin{tabular}{xxxxxxx}
    \small{Source} &
    \small{Target} &
    \small{Swap} & 
    \small{Coarse Only} & 
    \small{Middle Only} & 
    \small{Fine1 Only} & 
    \small{Fine2 Only} 
    \\
    \multicolumn{7}{c}{
    \includegraphics[width=\columnwidth]{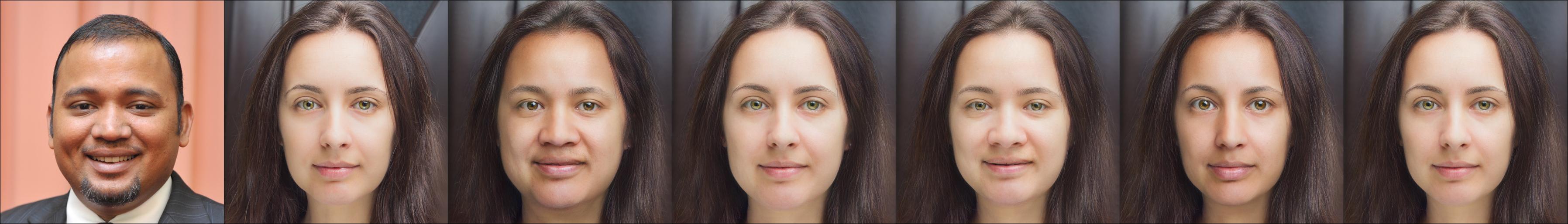}
    }
    \\[-2pt]
    \multicolumn{7}{c}{
    \includegraphics[width=\columnwidth]{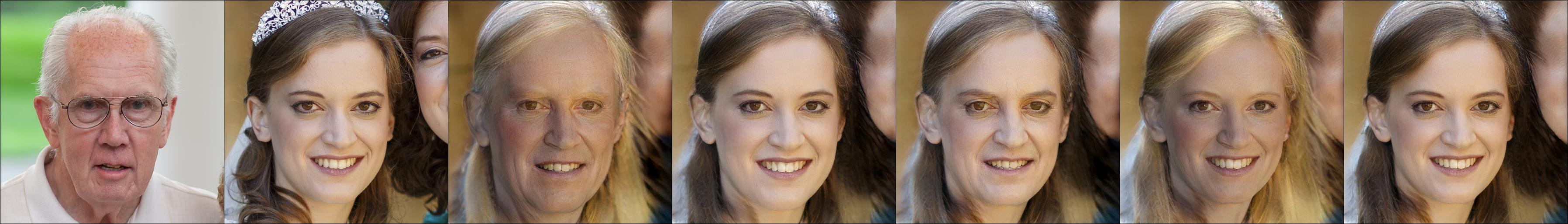}
    }
    \\[-2pt]
    \multicolumn{7}{c}{
    \includegraphics[width=\columnwidth]{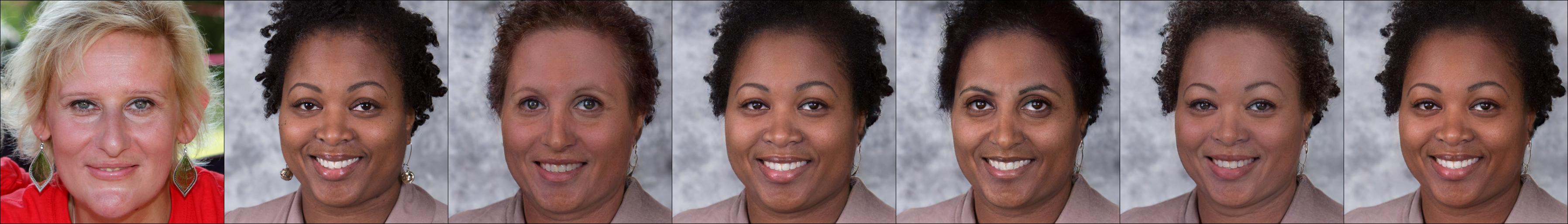}
    }
    \end{tabular}
\caption{\textbf{Resolution-wise analysis on face swapping results.} For each case, we use the swapped latent code \textit{only} at the respective resolution noted, while using target latent code otherwise. 
}
\label{fig:layerwise}
\vspace{-4mm}
\end{figure}

We classify the $18$ layers of the $\mathcal{W+}$ space by resolution: coarse ($4^2$ -- $8^2$), middle ($16^2$ -- $32^2$), fine$1$ ($64^2$ -- $128^2$), and fine$2$ ($256^2$ -- $1024^2$), and investigate the face swap results when only specific layers use the latent code of $l_{\text{swap}}$ while using the baseline $l_{\text{t}}$ for others. Figure~\ref{fig:layerwise} shows the results.

Results show that specific attributes are associated with particular resolutions in the $\mathcal{W+}$ space, and displays a direct correlation with the size of the receptive field of the resolutions. More specifically, coarse resolutions control attributes such as glasses or face structure. Middle resolutions are the most important in swapping as they contain the most information regarding the source identity. Fine1 resolutions embed information on smaller characteristic length attributes such as source skin, hair, and eye colors. We find that the fine$2$ resolutions, in general did not contain information on distinguishable attributes.

\begin{figure}[t]
\centering
    \setlength{\tabcolsep}{0pt}
    \newcolumntype{x}{>{\centering\arraybackslash\vspace{0pt}}m{0.17\linewidth}}
    \begin{tabular}{xxxxxx}
    \footnotesize{Source} &
    \footnotesize{Target} & 
    \footnotesize{Swap} &
    \footnotesize{Source} &
    \footnotesize{Target} & 
    \footnotesize{Swap}
    \\
    \multicolumn{6}{c}{
    \includegraphics[width=0.5\columnwidth]{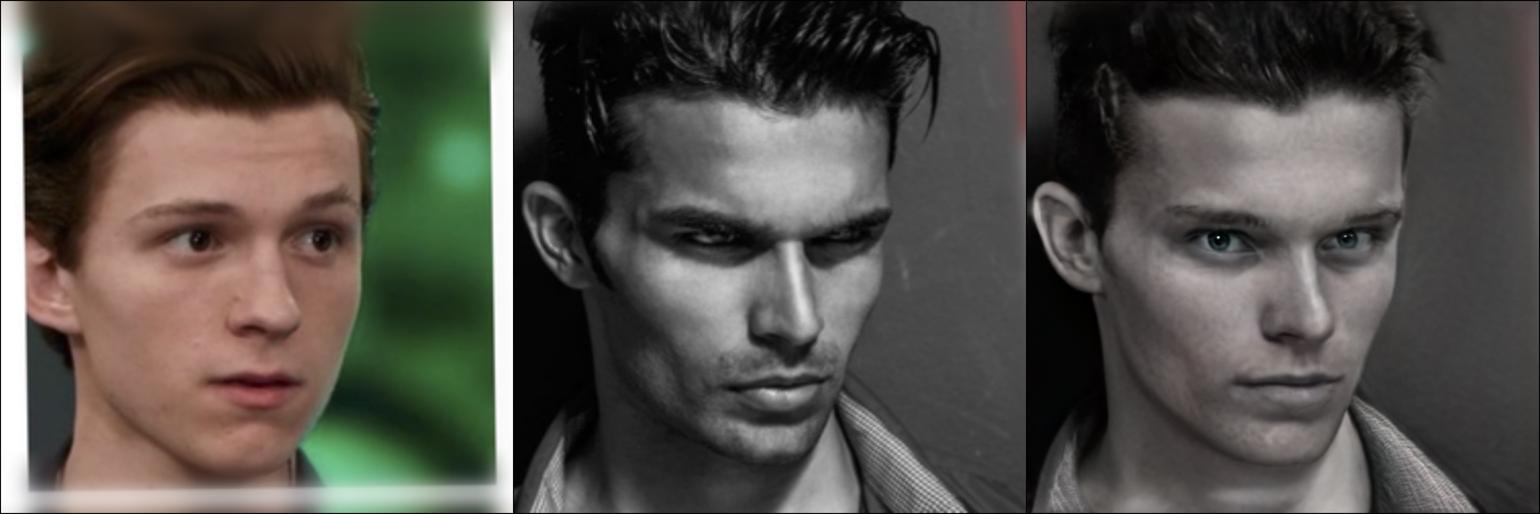}
    \includegraphics[width=0.5\columnwidth]{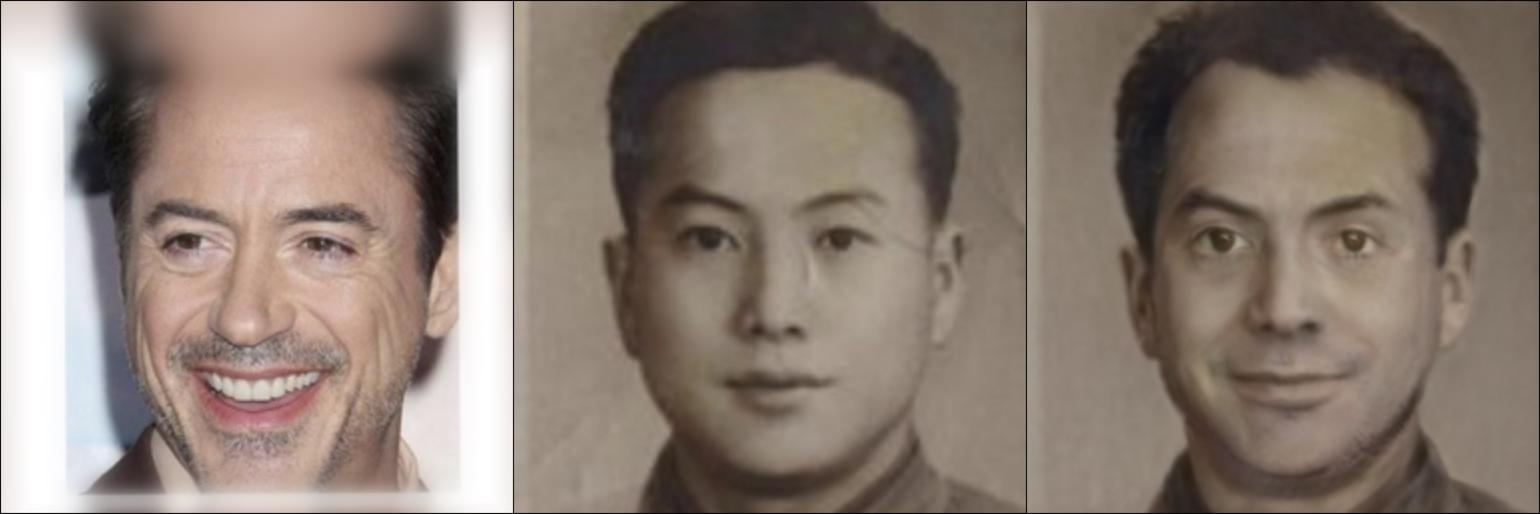}
    }
    \\[-2pt]
    \multicolumn{6}{c}{
    \includegraphics[width=0.5\columnwidth]{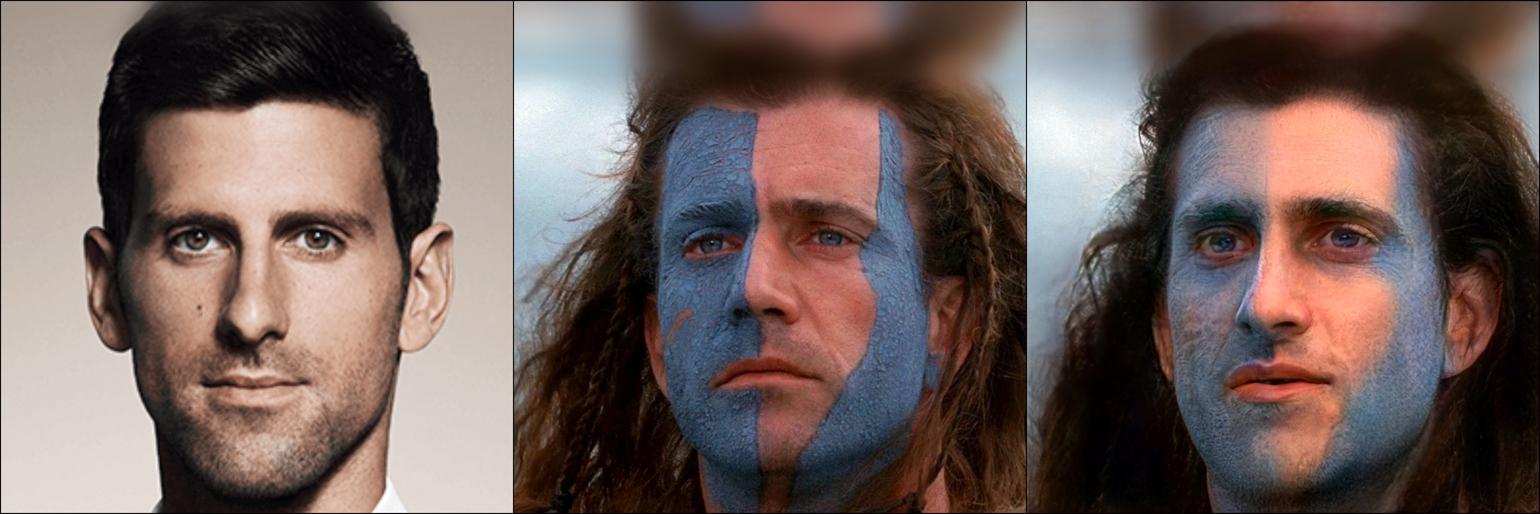}
    \includegraphics[width=0.5\columnwidth]{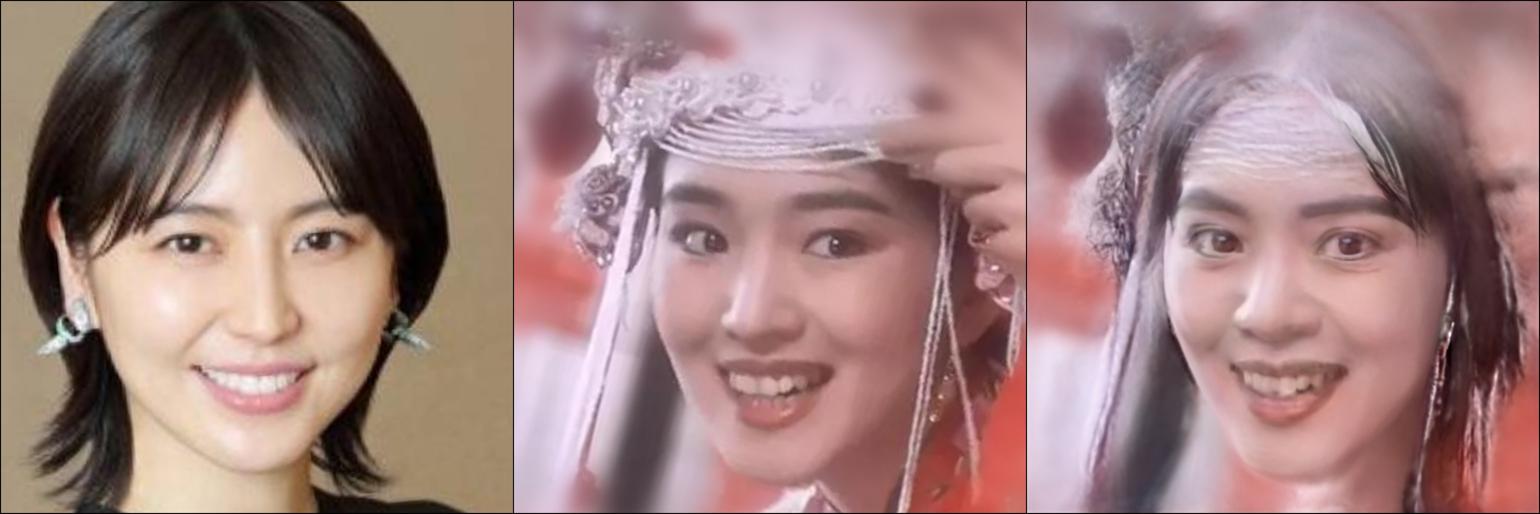}
    }
    \\
    \end{tabular}
\caption{\textbf{Face swapping results on wild images}. Each source-target pairs are extracted from the generic photo rather than a human face dataset.}
\label{fig:wild}
\vspace{-4mm}
\end{figure}

\vspace{3pt}\noindent\textbf{Wild Images.} Finally, we discuss our framework on wild images instead of human face specific datasets to test the robustness of our model on extreme cases, shown in Figure~\ref{fig:wild}.

After carefully tuning the appropriate loss objectives, the results demonstrate that the model is capable of producing convincing face swap results on grayscale images, or in multiple-color faces and images with complex backgrounds.


\begin{figure}[t]
  \includestandalone[width=0.47\textwidth]{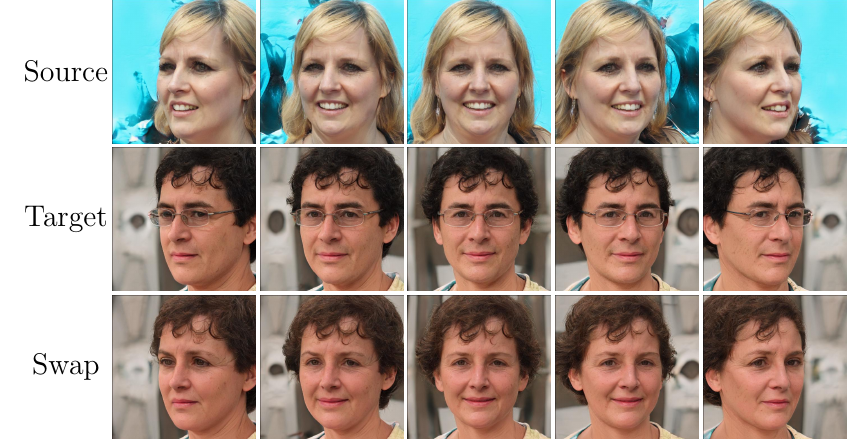}
  \caption{\textbf{3D-Aware face swapping results via StyleNeRF.} Images were generated for multiple different camera parameters. Face swap is maintained throughout different angles displaying good quality overall.
}
  \label{fig:stylenerf}
  \vspace{-4mm}
\end{figure}

\subsection{LatentSwap for 3D-Aware Generator} 
We apply the LatentSwap framework to the StyleNeRF~\cite{gu2021stylenerf} generator instead of StyleGAN2 generator to demonstrate that our framework can be applied to generate effective swapped images in a 3D-Aware multiview context.

We maintain the same training and sampling process to create the $\mathcal{W}$ space latent codes for the StyleNeRF generator, with the only difference is mapping it to the $21 \times 512$ dimensional latent code compatible with the StyleNeRF generator (unlike StyleGAN2 $\mathcal{W+}$ space's $18 \times 512$). 
The swapped results are shown in Figure~\ref{fig:stylenerf}

Our simple framework applies to the StyleNeRF generator, and we achieve good quality face swap for all different orientations of the resultant face images. This shows that our framework is applicable to a 3D-based model and opens up a wide array of opportunities to explore face swapping further in this direction in future works.

\subsection{Application: Latent Space Manipulation} 
\begin{figure}[t]
\centering
    \setlength{\tabcolsep}{0pt}
    \newcolumntype{x}{>{\centering\arraybackslash\vspace{0pt}}m{0.11\linewidth}}
    \begin{tabular}{xxxxxxxxx}
    \scriptsize{Source} &
    \scriptsize{Target} &
    \scriptsize{Swap} & 
    \scriptsize{Age+} & 
    \scriptsize{Age-} & 
    \scriptsize{Smile+} & 
    \scriptsize{Smile-} & 
    \scriptsize{Pose+} & 
    \scriptsize{Pose-}
    \\
    \multicolumn{9}{c}{
    \includegraphics[width=\columnwidth]{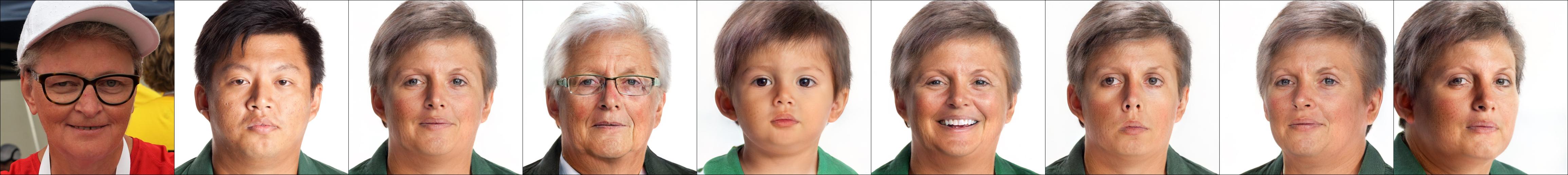}
    }
    \\[-2.5pt]
    \multicolumn{9}{c}{
    \includegraphics[width=\columnwidth]{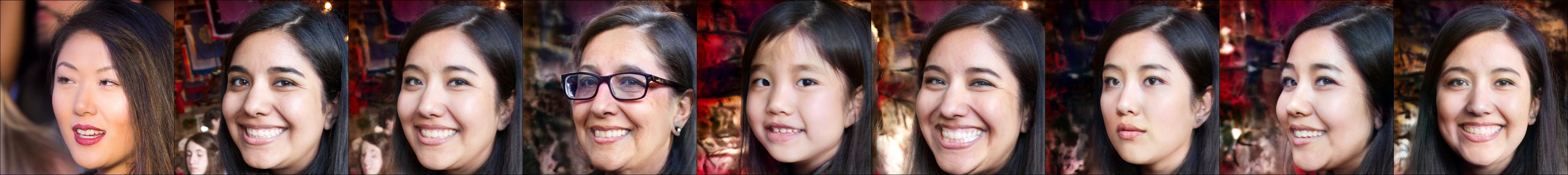}
    }
    \\[-2.5pt]
    \multicolumn{9}{c}{
    \includegraphics[width=\columnwidth]{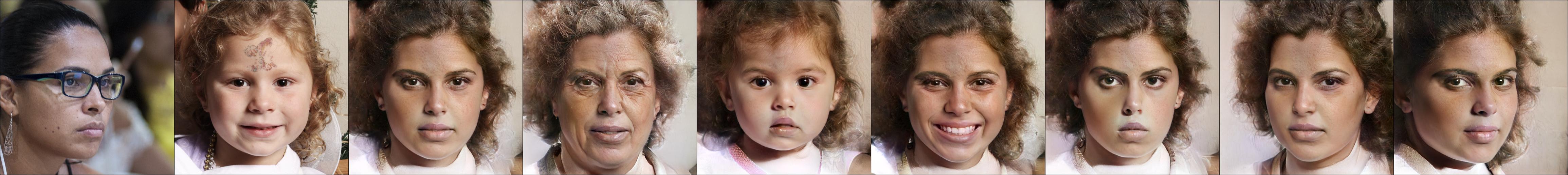}
    }
    \end{tabular}
\caption{\textbf{Editing results on the face swapped images for various attributes.} Using InterfaceGAN, the resultant images successfully edits age, smile, and pose, without losing the source identity.}
\label{fig:interfaceGAN}
\vspace{-6mm}
\end{figure}

We discuss briefly on editing specific attributes (e.g. pose, facial expression) by just using the latent code without the need of further inputs, which is possible as our model do not requiring additional modules. We apply InterFaceGAN~\cite{shen_2020} as the image editing module. From Figure~\ref{fig:interfaceGAN} we can verify that InterFaceGAN is able to edit various specific attributes such as age, smile, or pose of the swapped faces qualitatively well. 

Using only latent codes for editing implies any method editing images on the StyleGAN2 domain will be applicable to our result. We give the result for GANSpace~\cite{harkonen_2020} in Appendix 3 as an additional useful example.

\section{Conclusions}
\label{sec:conclusions}

In this paper, we propose LatentSwap, a simple fully-connected layer based model for face swapping in the StyleGAN~\cite{karras_2019, karras_2020_stylegan2, karras_2021} domain. Our model needs only randomly sampled latent codes for training and can perform good quality face swapping on arbitrary images given a pre-trained inversion network and a pre-trained generator. Our approach enjoys fast and stable training, benefiting from the stability of the pre-trained models and the simplicity of the framework. The face swapping outputs are realistic and have high resolution. In addition, our results are controllable between source and target images by altering the coefficient between the loss terms. We analyzed the role of each spatial resolutions of the swapped image latent code, and demonstrated that our framework is applicable to other generators such as StyleNeRF~\cite{gu2021stylenerf} and can work in conjunction with other StyleGAN2~\cite{karras_2020_stylegan2} downstream tasks.


\bibliography{anonymous-submission-latex-2024}

\end{document}